\documentclass[submission,copyright,creativecommons]{eptcs}
\usepackage{mathptmx}
\usepackage{amsmath}
\usepackage{amsfonts}
\usepackage{amssymb}
\usepackage{graphicx}%
\providecommand{\U}[1]{\protect\rule{.1in}{.1in}}

\begin{document}

\title{Splitting a Hybrid ASP Program}
\author{Alex Brik\\Google Inc., USA}
\def\titlerunning{Splitting a Hybrid ASP Program}
\def\authorrunning{A. Brik}
\maketitle

\begin{abstract}
Hybrid Answer Set Programming (Hybrid ASP)\ is an extension of Answer Set
Programming (ASP)\ that allows ASP-like rules to interact with outside
sources. The Splitting Set Theorem is an important and extensively used result
for ASP. The paper introduces the Splitting Set Theorem for Hybrid ASP, which
is for Hybrid ASP the equivalent of the Splitting Set Theorem, and shows how
it can be applied to simplify computing answer sets for Hybrid ASP programs
most relevant for practical applications.

\end{abstract}

\bigskip An important result for logic programs is the Splitting Set Theorem
\cite{DBLP:conf/iclp/LifschitzT94}, which shows how computing an answer set
for a program can be broken into several tasks of the same kind for smaller
programs. The theorem and its more general variant the Splitting Sequence
Theorem are extensively used for proving other theorems, for instance in
\cite{BalducciniG03}, \cite{DworschakGNSS08} or
\cite{DBLP:journals/ai/Bonatti08} among many others. Hybrid Answer Set
Programming (Hybrid ASP) \cite{BrikR11} is an extension of ASP\ that allows
ASP-like rules to interact with outside sources, which makes Hybrid ASP well
suited for practical applications. For instance, recently Hybrid ASP\ has been
used in a system for diagnosing failures of data processing pipelines at
Google Inc \cite{BrikX20}. The theory of Hybrid ASP, however\ is not
extensively developed. This paper introduces the Splitting Set Theorem for
Hybrid ASP and the Splitting Sequence Theorem for Hybrid ASP, which are the
equivalents for Hybrid ASP\ of the similarly named results for ASP, thus
making a small step towards developing the theory of Hybrid ASP. The author
hopes that the new theorems will have many future applications, in the way
analogous to the original Splitting Set Theorem and Splitting Sequence
Theorem. The potential of the new theorems to be useful in the future, and the
significance of the new results is demonstrated by using them to simplify
computation of answer sets for the types of Hybrid ASP\ programs most relevant
for practical applications, i.e. those applications that have answer sets with states having
times of the form $k\cdot\Delta t$, such as the programs that result from
translating descriptions in action languages Hybrid AL \cite{BrikR17} and
Hybrid ALE \cite{BomansonB19}, or such as the programs used in other
applications of Hybrid ASP\ \cite{BrikR15}, \cite{HybridASP2}.\  

The paper is structured as follows. The first section reviews ASP, The
Splitting Set Theorem and Hybrid ASP. The paper then presents The Splitting
Set Theorem for Hybrid ASP\ and The Splitting Sequence Theorem for Hybrid ASP.
The following section presents an algorithm that simplifies computing answer
sets for Hybrid ASP. Finally a short conclusion follows.

\section{Review of\ the Splitting Set Theorem and Hybrid ASP}

We will begin with a brief review of ASP. Let $At$ be a nonempty set of
symbols called \emph{atoms}.  A \emph{block} is an expression of the form
\begin{equation}
b_{1},...,b_{k},\;not\;b_{k+1},\;...,\;not\;b_{k+m}\label{block}%
\end{equation}
where $b_{1},\;...,\;b_{k+m}$ are atoms. For a block $B$ as above, let the
\emph{set of atoms of }$B$ be defined as $At\left(  B\right)  \equiv
\{b_{1},\;...,\;b_{k+m}\}$. $B^{+}\equiv b_{1},\;...,\;b_{k}$ is called the
\emph{positive part of }$B$, and $B^{-}\equiv$ $not\;b_{k+1}%
,\;...,\;not\;b_{k+m}$ is called the \emph{negative part of }$B$. A set
operation applied to a block $B$ will indicate the same set operation applied
to $At\left(  B\right)  $ with the block being reconstructed from the result
of the set operation. For instance $b_{1},\;b_{2},\;not\;b_{3},\;b_{4}%
\;\backslash\;\{b_{1},\;b_{4}\}$ will indicate a block $b_{2},\;not\;b_{3}$.

A \emph{normal propositional logic programming }rule is an expression of the
form%
\begin{equation}
r\equiv a:-\;B\label{clause2}%
\end{equation}
where $a$ is an atom and $B$ is a block. We define the \emph{head of }$r$ as
$head\left(  r\right)  \equiv a$, and we define the \emph{body of }$r$ as
$body\left(  r\right)  \equiv B$. We define $At\left(  r\right)
\equiv\left\{  a\right\}  \cup At\left(  B\right)  $.

\bigskip Given any set $M\subseteq At$ and a block $B$, we say that $M$
satisfies $B$, written $M\models B$, if $At\left(  B^{+}\right)  \subseteq M$
and $At\left(  B^{-}\right)  \cap M=\emptyset$. For a rule $r$, we say that
$M$ satisfies $r$, written $M\models r$, if whenever $M$ satisfies the body of
$r$, then $M$ satisfies the head of $r$. A \emph{normal logic program} $P$ is
a set of rules. We say that $M\subseteq At$ is a model of $P$, written
$M\models P$, if $M$ satisfies every rule of $P$.

A \emph{Horn rule} is the rule with the empty negative part. A Horn program
$P$ is a set of Horn rules. Each Horn program $P$ has a least model under
inclusion, $LM_{P}$, which can be defined using the \emph{one-step provability
operator} $T\left[  P\right]  $ as follows. For any set $A$, let
$\mathcal{P}\left(  A\right)  $ denote the set of all subsets of $A$. The
one-step provability operator $T\left[  P\right]  :\mathcal{P}\left(
At\right)  \rightarrow\mathcal{P}\left(  At\right)  $ associated with the Horn
program $P$ \cite{EmdenK76} is defined by setting
\[
T\left[  P\right]  (M)=M\cup\{a:\exists r\in P\;(a=head(r)\wedge M\models
body(r))\}
\]
for any $M\in\mathcal{P}\left(  At\right)  $. We define $T\left[  P\right]
^{n}(M)$ by induction by setting $T\left[  P\right]  ^{0}\left(  M\right)
=M$, $T\left[  P\right]  ^{1}(M)=T\left[  P\right]  (M)$ and $T\left[
P\right]  ^{n+1}(M)=T\left[  P\right]  (T\left[  P\right]  ^{n}(M))$. Then the
least model $LM_{P}$ can be computed as \linebreak $LM_{P}=\bigcup_{n\geq0}T\left[
P\right]  ^{n}(\emptyset).$

If $P$ is a normal logic program and $M\subseteq At$, then the
Gelfond-Lifschitz (GL)\ reduct of $P$ with respect to $M$ \cite{GelfondL88} is
the Horn program $P^{M}$ which results by eliminating those rules $r$ such
that $M\not \models body\left(  r\right)  ^{-}$ and replacing other rules $r$
by $head\left(  r\right)  :-body\left(  r\right)  ^{+}$. We then say that $M$
is a \emph{stable model} for $P$ if $M$ equals the least model of $P^{M}$.

An \emph{answer set programming rule }is an expression of the form
(\ref{clause2}) where $a,b_{1},\ldots,b_{k+m}$ are classical literals, i.e.,
either positive atoms or atoms preceded by the classical negation sign $\lnot
$. The set of literals of $At$ will be denoted $Lit_{At}$. \emph{Answer sets}
are defined in analogy to stable models, but taking into account that atoms
may be preceded by classical negation and that atoms $a$ and classically
negated atoms $\lnot a$ are mutually exclusive in answer sets.

We will now follow \cite{DBLP:conf/iclp/LifschitzT94} in review of the
Splitting Set Theorem and the Splitting Sequence Theorem. A \emph{splitting
set} for a program $P$ is any set $U\subseteq At$ such that for every rule
$r\in P$ if $head\left(  r\right)  \in U$ then $At\left(  r\right)  \subseteq
U$. The set of rules $r\in P$ such that $At\left(  r\right)  \subseteq U$ is
called the \emph{bottom }of $P$ relative to the splitting set $U$ and is
denoted by $b_{U}\left(  P\right)  $. The set $P\backslash b_{U}\left(
P\right)  $ is the \emph{top} of $P$ relative to $U$. 

Consider $X\subseteq At$. For each rule $r\in P$ such that $At(body\left(
r\right)  ^{+})\cap U\subseteq X$ and $At(body\left(  r\right)  ^{-})\cap
U\cap X=\emptyset$ take the rule $r^{\prime}$ defined by%
\[
head\left(  r\right)  \;:-\;body\left(  r\right)  \;\backslash\;U
\]

The program consisting of all rules $r^{\prime}$ obtained in this way will be
denoted by $\epsilon_{U}\left(  P,X\right)  $.

A \emph{solution }to $P$ with respect to $U$ is a pair $\left(  X,Y\right)  $
of sets of literals such that

\begin{itemize}
\item $X$ is an answer set for $b_{U}\left(  P\right)  $

\item $Y$ is an answer set for $\epsilon_{U}\left(  P\;\backslash
\;b_{U}\left(  P\right)  ,\;X\right)  $

\item $X\cup Y$ is consistent (a set is consistent if for any atom $a$ it does
not contain both $a$ and classically negated atom $-a$)
\end{itemize}

\textbf{Splitting Set Theorem. }\emph{Let }$U$\emph{ be a splitting set for a
program }$P$\emph{. A set }$A$\emph{ of literals is a consistent answer set
for }$P$\emph{ if and only if }$A=X\cup Y$\emph{ for some solution }$\left(
X,Y\right)  $\emph{ to }$P$\emph{ with respect to }$U$\emph{.}

\bigskip

We will now review extending the definition of a splitting set to a splitting
sequence. A \emph{sequence} is a family whose index set is an initial segment
of ordinals, $\{\alpha:\alpha<\mu\}$. The ordinal $\mu$ is the \emph{length
}of the sequence. A sequence $\left\langle U_{\alpha}\right\rangle
_{\alpha<\mu}$ of sets is \emph{monotone} if $U_{\alpha}\subset U_{\beta}$
whenever $\alpha<\beta$, and \emph{continuous }if, for each limit ordinal
$\alpha<\mu$, $U_{\alpha}=%
{\displaystyle\bigcup\limits_{\beta<\alpha}}
U_{\beta}$.

A \emph{splitting sequence }for a program $P$ is a monotone, continuous
sequence $\left\langle U_{\alpha}\right\rangle _{\alpha<\mu}$ of splitting
sets for $P$ such that $%
{\displaystyle\bigcup\limits_{\alpha<\mu}}
U_{\alpha}=Lit_{At}$. The definition of a solution with respect to a splitting
set is extended to splitting sequence as follows. A \emph{solution} to $P$
with respect to $\left\langle U_{\alpha}\right\rangle _{\alpha<\mu}$ is a
sequence $\left\langle X_{\alpha}\right\rangle _{\alpha<\mu}$ of sets of
literals such that

\begin{itemize}
\item $X_{0}$ is an answer set for $b_{U_{0}}\left(  P\right)  $,

\item for any $\alpha$ such that $\alpha+1<\mu$, $X_{\alpha+1}$ is an answer
set for $\epsilon_{U_{\alpha}}(b_{U_{\alpha+1}}\left(  P\right)  \backslash
b_{U_{\alpha}}\left(  P\right)  ,\;%
{\displaystyle\bigcup\limits_{\beta\leq\alpha}}
X_{\beta})$,

\item for any limit ordinal $\alpha<\mu$, $X_{\alpha}=\emptyset$,

\item $%
{\displaystyle\bigcup\limits_{\alpha<\mu}}
X_{\alpha}$ is consistent.
\end{itemize}

\textbf{Splitting Sequence Theorem.}\emph{ Let }$U\equiv\left\langle
U_{\alpha}\right\rangle _{\alpha<\mu}$\emph{ be a splitting sequence for a
program }$P$\emph{. A set }$A$\emph{ of literals is a consistent answer set
for }$P$\emph{ if and only if }$A=%
{\displaystyle\bigcup\limits_{\alpha<\mu}}
X_{\alpha}$\emph{ for some solution }$\left\langle X_{\alpha}\right\rangle
_{\alpha<\mu}$\emph{ to }$P$\emph{ with respect to }$U$\emph{.}

\bigskip

We will now proceed with the review of Hybrid ASP. A Hybrid ASP\ program $P$
has an underlying parameter space $S$. Elements of $S$ are of the form
$\mathbf{p}=(t,x_{1},\ldots,x_{l})$ where $t$ is time and $x_{i}$ are
arbitrary parameter values. We shall let $t(\mathbf{p})$ denote $t$ and
$x_{i}(\mathbf{p})$ denote $x_{i}$ for $i=1,\ldots,l$. We refer to the
elements of $S$ as \emph{generalized positions}. Let $At$ be a set of atoms of
$P$. Then the universe of $P$ is $At\times S$. Let $B$ be a block. We will
define%
\[
B\times\mathbf{p}\equiv\{\left(  x,\mathbf{p}\right)  :\;x\in B\}.
\]

If $M\subseteq At\times S$, we let $GP(M)=\{\mathbf{p}\in S:(\exists a\in
At)((a,\mathbf{p})\in M)\}$. Given an \emph{initial condition}, defined as a
subset  $I\subseteq S$ let $GP_{I}\left(  M\right)  =GP\left(  M\right)  \cup
I$. Given $M\subseteq At\times S$ and $\mathbf{p}\in S$, we say that $M$ and
initial condition $I$ satisfy a block $B$ of the form (\ref{block}) at the
generalized position $\mathbf{p}$, written $M\models_{I}\left(  B,\mathbf{p}%
\right)  $, if the following holds:

\begin{itemize}
\item if $B^{+}\neq\emptyset$ then $B^{+}\times\mathbf{p}\subseteq M$ and
$B^{-}\times\mathbf{p\cap M=\emptyset}$ 

\item if $B^{+}=\emptyset$ then $B^{-}\times\mathbf{p\cap M=\emptyset}$ and
$\mathbf{p}\in GP_{I}\left(  M\right)  $.
\end{itemize}

We say that $M$ satisfies a n-tuple of blocks written as $B_{1};\;...;\;B_{n}$
with the initial condition $I$ at the n-tuple of generalized positions
$(\mathbf{p}_{1},\;...,\;\mathbf{p}_{n})$, written $M\models_{I}%
(B_{1};\;...;B_{n},\;(\mathbf{p}_{1},\;...,\mathbf{p}_{n}))$, if $M\models
_{I}(B_{i},\mathbf{p}_{i})$ for $i=1,...,n$.

There are two types of rules in Hybrid ASP. \textbf{Advancing rules} are of
the form
\begin{equation}
r\equiv a:-B_{1};B_{2};\ldots;B_{n}:A,O\label{advrule}%
\end{equation}
where $A$ is a function returning a set of generalized positions, $body\left(  r\right)  \equiv B_{1},\;...,\;B_{n}$
are blocks, $head\left(  r\right)  \equiv a$ is a literal, and $O$ is a subset
of $S^{n}$ such that if $(\mathbf{p}_{1},\ldots,\mathbf{p}_{n})\in O$, then
$t(\mathbf{p}_{1})<\cdots<t(\mathbf{p}_{n})$ and $A\left(  \mathbf{p}%
_{1},\ldots,\mathbf{p}_{n}\right)  $ ($A$ applied to $ \mathbf{p}_{1},\ldots,\mathbf{p}_{n}$) is a subset of $S$ such that for all
$\mathbf{q}\in A\left(  \mathbf{p}_{1},\ldots,\mathbf{p}_{n}\right)  $,
$t(\mathbf{q})>t(\mathbf{p}_{n})$. Here and in the next rule, we allow blocks
to be empty for any $i$. $O$ is called the \emph{constraint set} of the rule
$r$ and will be denoted by $CS(r)$. $A$ is called the \emph{advancing
algorithm} of the rule $r$ and is denoted by $\operatorname*{Adv}(r)$. The
arity of rule $r$, $N\left(  r\right)  $, is equal to $n$.

The idea is that if $(\mathbf{p}_{1},\ldots,\mathbf{p}_{n})\in O$ and for each
$i$, $B_{i}$ is satisfied at the generalized position $\mathbf{p}_{i}$, then
the function $A$ can be applied to $(\mathbf{p}_{1},\ldots,\mathbf{p}_{n})$
to produce a set of generalized positions $O^{\prime}$ such that if
$\mathbf{q}\in O^{\prime}$, then $t(\mathbf{q})>t(\mathbf{p}_{n})$ and
$(a,\mathbf{q})$ holds. Thus advancing rules are like input-output devices in
that the function $A$ allows the user to derive possible successor
generalized positions as well as certain atoms $a$ which are to hold at such
positions. The advancing algorithm $A$ can access outside sources quite
arbitrarily in that it may involve functions for solving differential or
integral equations, solving a set of linear equations or linear programming
equations, solving an optimization problem, etc. (as for example in
\cite{HybridASP2}).

\textbf{Stationary rules} are of the form
\begin{equation}
r\equiv a:-B_{1};B_{2};\ldots;B_{n}:H,O\label{strule}%
\end{equation}
\newline where $body\left(  r\right)  \equiv B_{1},...,B_{n}$ are blocks,
$head\left(  r\right)  \equiv a$ is a literal, $H$ is called a \emph{boolean algorithm} of
the rule $r$ and will be denoted by $Bool\left(  r\right)  $, and $O\subseteq
S^{k}$ is the constraint set of the rule $r$ denoted $CS(r)$. A boolean algorithm is a function returning either true or false. We will
sometimes treat a boolean algorithm of the rule as a set. For instance $H\cap
O$ will indicate all the n-tuples of generalized positions $(\mathbf{p}%
_{1},\ldots,\mathbf{p}_{n})$ such that $H\left(  \mathbf{p}_{1},\ldots
,\mathbf{p}_{n}\right)  $ is true and $(\mathbf{p}_{1},\ldots,\mathbf{p}%
_{n})\in O$. The arity of rule $r$, $N\left(  r\right)  $, is equal to $n$.

Stationary rules are much like normal logic programming rules in that they
allow us to derive new atoms at a given generalized position $\mathbf{p}_{n}$.
The idea is that if $(\mathbf{p}_{1},\ldots,\mathbf{p}_{n})\in O\cap H$ and
for each $i$, $B_{i}$ is satisfied at the generalized position $\mathbf{p}%
_{i}$, then $(a,\mathbf{p}_{n})$ holds. The difference is that a derivation
with our stationary rules can depend on what happens in the multiple past time
points and the boolean algorithm $H$ can be any sort of a function which
returns either true or false. \ \newline

For an advancing rule or a stationary rule $r$ as above we define the\emph{
positive part of the body} of $r$, denoted $body\left(  r\right)  ^{+}\equiv
B_{1}^{+};...;B_{n}^{+}$ and we define the \emph{negative part of the body }of
$r$, denoted $body\left(  r\right)  ^{-}\equiv B_{1}^{-};...;B_{n}^{-}$. For
the rest of the paper, we denote by $n$ the arity of a hybrid ASP\ rule when
the rule is clear from the context.

A Hybrid ASP program $P$ is a collection of Hybrid ASP advancing and
stationary rules. To define the notion of a stable model of $P$, we first must
define the notion of a Hybrid ASP Horn program and the one-step provability
operator for Hybrid ASP Horn programs.

A \emph{Hybrid ASP Horn program} is a Hybrid ASP program which does not
contain any negated atoms. Let $P$ be a Horn Hybrid ASP\ program and
$I\subseteq S$ be an initial condition. Then the one-step provability operator
$T\left[  P,I\right]  $ is defined so that given $M\subseteq At\times S$,
$T\left[  P,I\right]  (M)$ consists of $M$ together with the set of all
$(a,J)\in At\times S$ such that

\begin{enumerate}
\item \noindent there exists a stationary rule $r$ and $(\mathbf{p}_{1}%
,\ldots,\mathbf{p}_{n})\in CS\left(  r\right)  \cap Bool\left(  r\right)
\cap\left(  GP_{I}(M)\right)  ^{n}$ such that \linebreak$(head\left(  r\right)
,J)=(a,\mathbf{p}_{n})$ and $M\models(body\left(  r\right)  ,\;(\mathbf{p}%
_{1},...,\mathbf{p}_{n}))$ or

\item there exists an advancing rule $r$ and $(\mathbf{p}_{1},\ldots
,\mathbf{p}_{n})\in CS\left(  r\right)  \cap\left(  GP_{I}(M)\right)  ^{n}$
such that\linebreak\ $J\in\operatorname*{Adv}\left(  r\right)  (\mathbf{p}%
_{1},\ldots,\mathbf{p}_{n})$ and $M\models(body\left(  r\right)
,\;(\mathbf{p}_{1},...,\mathbf{p}_{n}))$ and $a=head\left(  r\right)  $.
\end{enumerate}

The stable model semantics for Hybrid ASP programs is defined as follows. Let
$M\subseteq At\times S$ and $I$ be an initial condition in $S$. An Hybrid ASP
rule $r\equiv a:-B_{1};\ldots,B_{n}:A,O$ is \emph{inapplicable} for $(M,I)$ if
for all $(\mathbf{p}_{1},\ldots,\mathbf{p}_{n})\in O\cap\left(  GP_{I}%
(M)\right)  ^{n}$, either (i) there is an $i$ such that $M\not \models
(B_{i}^{-},\mathbf{p}_{i})$, (ii) $A\left(  \mathbf{p}_{1},\ldots
,\mathbf{p}_{n}\right)  \cap GP_{I}(M)=\emptyset\,$ if $A$ is an advancing
algorithm, or (iii) $A(\mathbf{p}_{1},\ldots,\mathbf{p}_{n})=0$ if $A$ is a
boolean algorithm. 

\bigskip If $r$ is not inapplicable for $\left(  M,I\right)  $ then we define
the GL\ reduct of $r$ over $M$ and $I$, denoted by $r^{M,I}$ as follows:

\begin{enumerate}
\item If $r$ is an advancing rule $r\equiv a:-B_{1};...;B_{n}:A,O$ then
$r^{M,I}\equiv B_{1}^{+};\ldots,B_{n}^{+}:A^{M,I},O^{M,I}$ where $O^{M,I}$ is
equal to the set of $(\mathbf{p}_{1},\ldots,\mathbf{p}_{n})$ in $O\cap\left(
GP_{I}(M)\right)  ^{n}$ such that \ $M\models_{I}(body\left(
r\right)  ^{-},\;(\mathbf{p}_{1},\ldots,\mathbf{p}_{n}))$ and $A(\mathbf{p}%
_{1},\ldots,\mathbf{p}_{n})\cap GP_{I}(M)\neq\emptyset$, and $A^{M,I}%
(\mathbf{p}_{1},\ldots,\mathbf{p}_{n})\equiv A(\mathbf{p}_{1},\ldots
,\mathbf{p}_{n})\cap GP_{I}(M)$.

\item If $r$ is a stationary rule $r\equiv a:-B_{1};...;B_{n}:A,O$ then
$r^{M,I}\equiv a:-B_{1}^{+};\ldots,B_{n}^{+}:H|_{O^{M,I}},O^{M,I}$ where
$O^{M,I}$ is equal to the set of all $(\mathbf{p}_{1},\ldots,\mathbf{p}_{n})$
in $O\cap\left(  GP_{I}(M)\right)  ^{n}$ such that \linebreak $\ M\models
_{I}(body\left(  r\right)  ^{-},\;(\mathbf{p}_{1},\ldots,\mathbf{p}_{n}))$ and
$H(\mathbf{p}_{1},\ldots,\mathbf{p}_{n})$ is true.
\end{enumerate}

\bigskip One note to make about the definition above is that GL\ reduct cannot
derive generalized positions that are not in $GP_{I}\left(  M\right)  $. This
is because the range of $A^{M,I}$ in the definition is restricted to
$GP_{I}(M)$.

We form a GL reduct of $P$ over $M$ and $I$, $P^{M,I}$ as follows.

\begin{enumerate}
\item \noindent Eliminate all rules which are inapplicable for $(M,I)$.

\item If a rule $r\in P$ is not eliminated in step 1, then replace it by the
rule $r^{M,I}$.
\end{enumerate}

We then say that $M$ is a\emph{ stable model of }$P$\emph{\ with initial
condition }$I$ if $%
{\displaystyle\bigcup\limits_{k=0}^{\infty}}
T\left[  P^{M,I},I\right]  ^{k}\left(  \emptyset\right)  =M.$

\emph{Answer sets} are defined in analogy to stable models, but taking into
account that atoms may be preceded by classical negation and that
$(a,\mathbf{p)}$ and $\left(  -a,\mathbf{p}\right)  $ are mutually exclusive
in answer sets.

\section{The Splitting Set Theorem for Hybrid ASP}

We will now introduce additional notation that will be used
throughout the rest of the paper.

Without loss of generality assume that all advancing rules are of the form%
\[
a:-B_{1};\;...;\;B_{n}:O,A
\]
and all of stationary rules are of the form%
\[
a:-B_{1};\;...;\;B_{n}:O,H
\]
where $a$ is a literal, $B_{1}$, ..., $B_{n}$ are blocks, $O$ is a constraint
set, $A$ is an advancing algorithm, and $H$ is a boolean algorithm.

Let $M$ be a set of literals and generalized position pairs, and let
$\mathbf{p}$ be a generalized position. Define%
\[
M|_{\mathbf{p}}\equiv\{\left(  a,\mathbf{q}\right)  \in M\;:\;\mathbf{q=p}\}
\]%
\[
At\left(  M\right)  \equiv\{a\;:\;\left(  a,\mathbf{p}\right)  \in M\}
\]

Let $U\subseteq Lit_{At}\times S$. We say that $U$ is a \emph{splitting set of
}$P$ \emph{with initial condition} (w.i.c.) $J$ if for all $r\in P$ 

\begin{enumerate}
\item if $r$ is advancing and $(\mathbf{p}_{1},\;...,\;\mathbf{p}_{n})\in
CS\left(  r\right)  $ and $\mathbf{p}\in Adv\left(  r\right)  \left(
\mathbf{p}_{1},\;...,\;\mathbf{p}_{n}\right)  $ and $\left(  a,\mathbf{p}%
\right)  \in U$ then both for $i=1,...,n$, $B_{i}\times\mathbf{p}_{i}\subseteq
U$ and $\{\mathbf{p}_{1},\;...,\;\mathbf{p}_{n}\}\subseteq GP_{J}\left(
U\right)  $.

\item if $r$ is stationary and $(\mathbf{p}_{1},\;...,\;\mathbf{p}_{n})\in
CS\left(  r\right)  $ and $\left(  a,\mathbf{p}_{n}\right)  \in U$ then both
for $i=1,...,n$, $B_{i}\times\mathbf{p}_{i}\subseteq U$ and $\{\mathbf{p}%
_{1},\;...,\;\mathbf{p}_{n}\}\subseteq GP_{J}\left(  U\right)  $.
\end{enumerate}

\bigskip As in the case of the original splitting set theorem
\cite{DBLP:conf/iclp/LifschitzT94} the splitting set $U$ acts to split Hybrid
ASP\ program $P$ into the part that can derive $U$ or one of its subsets, and
the remaining part of $P$, which can derive $At\times S\backslash U$ or one of
its subsets. The difference, however, is that for a given rule the conclusion
of the rule may be in $U$ for some n-tuples of generalized positions
($\mathbf{p}_{1},\;...,\;\mathbf{p}_{n}$) and not for others. So, the
splitting set splits not only the program, but the rules themselves. This will
be elaborated below.

As in the case of the original splitting set theorem we identify by
$b_{U}\left(  P\right)  $ a set of new rules that capture the rules and
generalized positions that may contribute to generating $U$.

Define $Rules_{b}\left(  U,P\right)  $ as%
\[
\{\;r\in P\;:\;\text{if }r\text{ is advancing and there exists }%
(\mathbf{p}_{1},\;...,\;\mathbf{p}_{n})\in CS\left(  r\right)
\]%
\[
\text{and }\mathbf{p}\in Adv\left(  r\right)  (\mathbf{p}_{1}%
,\;...,\;\mathbf{p}_{n})\text{ such that }\left(  a,\mathbf{p}\right)  \in U
\]%
\[
\text{if }r\text{ is stationary and there exists }(\mathbf{p}_{1}%
,\;...,\;\mathbf{p}_{n})\in CS\left(  r\right)  \cap Bool\left(  r\right)
\]%
\[
\text{such that }\left(  a,\mathbf{p}_{n}\right)  \in U\;\}
\]

In other words, $Rules_{b}\left(  U,P\right)  $ is the set of all rules of $P$
that could contribute to $U$ for some tuple of generalized positions.

For an advancing rule $r$ let%
\[
CS_{b}\left(  U,r\right)  \equiv\{\;(\mathbf{p}_{1},\;...,\;\mathbf{p}_{n})\in
CS\left(  r\right)  \;:
\]%
\[
\text{there exists }p\in Adv\left(  r\right)  (\mathbf{p}_{1}%
,\;...,\;\mathbf{p}_{n})\text{ such that }\left(  a,\mathbf{p}\right)  \in
U\;\}
\]

For a stationary rule $r$ let%
\[
CS_{b}\left(  U,r\right)  \equiv\{\;(\mathbf{p}_{1},\;...,\;\mathbf{p}_{n})\in
CS\left(  r\right)  \cap Bool\left(  r\right)  \;:\;\left(  a,\mathbf{p}%
_{n}\right)  \in U\;\}
\]

That is $CS_{b}\left(  U,r\right)  $ are all the generalized position tuples
for which $r$ could contribute to $U$.

For an advancing rule $r\in Rules_{b}\left(  U,P\right)  $ define
$Adv_{b}\left(  U,r\right)  $ by%
\[
Adv_{b}\left(  U,r\right)  (\mathbf{p}_{1},\;...,\;\mathbf{p}_{n}%
)\equiv\{\;\mathbf{p}\;:\ \mathbf{p}\in Adv\left(  r\right)  (\mathbf{p}%
_{1},\;...,\;\mathbf{p}_{n})
\]%
\[
\text{such that }\left(  a,\mathbf{p}\right)  \in U\text{ if }(\mathbf{p}%
_{1},\;...,\;\mathbf{p}_{n})\in CS_{b}\left(  U,r\right)  \;\}
\]

$Adv_{b}\left(  U,r\right)  $ is an advancing algorithm that for any tuple of
generalized positions will only generate those $\mathbf{p}$ that contribute to
$U$.

For an advancing rule $r$ let%
\[
b_{U}\left(  r\right)  \equiv head\left(  r\right)  :-\;body\left(  r\right)
:CS_{b}\left(  U,r\right)  ,\;Adv_{b}\left(  U,r\right)
\]

For a stationary rule $r$ let%
\[
b_{U}\left(  r\right)  \equiv head\left(  r\right)  :-\;body\left(  r\right)
:CS_{b}\left(  U,r\right)  ,\;Bool\left(  r\right)
\]

Define the \emph{bottom of }$P$ with respect to $U$, $b_{U}\left(  P\right)  $
as%
\[
b_{U}\left(  P\right)  \equiv\{\;b_{U}\left(  r\right)  \;:\;r\in
Rules_{b}\left(  U,P\right)  \;\}
\]
The idea is that just like in \cite{DBLP:conf/iclp/LifschitzT94},
$b_{U}\left(  P\right)  $ forms only those rules that could contribute to $U$,
and so $X$ will be an answer set of $b_{U}\left(  P\right)  $ w.i.c. $J$ iff
$M\cap U=X$ for some answer set $M$ of $P$ w.i.c. $J$.

\bigskip

We will now proceed to define $\epsilon_{U}\left(  P,X\right)  $ with the
understanding that the same rule may contribute to $U$ for some generalized
position tuples and contribute to $Lit_{At}\times S\;\backslash U$ for others.

First, we need to identify remainder $\operatorname*{Rem}\left(  U,P\right)  $
of $Rules_{b}\left(  U,P\right)  $ not captured by $b_{U}\left(  P\right)  $.
That is we need to identify the parts contributing to the complement of $U$ of
those rules that have other parts contributing to $U$. This is due to an
important difference between Hybrid ASP\ and ASP. In ASP a rule contributes a
single conclusion. Thus if ASP rule contributes to the splitting set then it
must be in the bottom of the program. In Hybrid ASP, however, a rule acts more
like a collection of rules contributing different conclusions for different
generalized position tuples. Consequently, the parts of the rules that
contribute to the complement of the splitting set need to be separated from
those that contribute to the splitting set itself. We will now proceed with
the definition.

For an advancing rule $r$ define%
\[
CS_{\operatorname*{Rem}}\left(  U,r\right)  \equiv\{\ (\mathbf{p}%
_{1},\;...,\;\mathbf{p}_{n})\in CS\left(  r\right)  :
\]%
\[
\text{there exists }\mathbf{p}\in Adv\left(  r\right)  (\mathbf{p}%
_{1},\;...,\;\mathbf{p}_{n})\text{ }\left(  a,\mathbf{p}\right)  \notin U\;\}
\]

For a stationary $r$ define%
\[
CS_{\operatorname*{Rem}}\left(  U,r\right)  \equiv\{\;(\mathbf{p}%
_{1},\;...,\;\mathbf{p}_{n})\in CS\left(  r\right)  \cap Bool\left(  r\right)
:\;\left(  a,\mathbf{p}_{n}\right)  \notin U\;\}
\]

That is, $CS_{\operatorname*{Rem}}\left(  U,r\right)  $ contains those
generalized position tuples such that for them the rule $r$ contributes to the
complement of $U$.

For an advancing rule $r\in Rules_{b}\left(  U,P\right)  $ and $(\mathbf{p}%
_{1},\;...,\;\mathbf{p}_{n})$ define
\[
Adv_{\operatorname*{Rem}}(U,r)(\mathbf{p}_{1},\;...,\;\mathbf{p}_{n}%
)\equiv\left\{
\begin{array}
[c]{l}%
\{\mathbf{p}\;:\;\mathbf{p}\in Adv\left(  r\right)  (\mathbf{p}_{1}%
,\;...,\;\mathbf{p}_{n})\text{ s.t. }\left(  a,\mathbf{p}\right)  \notin
U\;\}\\
\text{ \ \ \ if }(\mathbf{p}_{1},\;...,\;\mathbf{p}_{n})\in
CS_{\operatorname*{Rem}}\left(  U,r\right)  \\
\emptyset\text{ if }(\mathbf{p}_{1},\;...,\;\mathbf{p}_{n})\notin
CS_{\operatorname*{Rem}}(U,r)
\end{array}
\right.
\]

\bigskip That is $Adv_{\operatorname*{Rem}}(U,r)$ is a restriction of
$Adv\left(  r\right)  $ to those generalized positions such that for them $r$
contributes to the complement of $U$.

When $CS_{\operatorname*{Rem}}\left(  U,r\right)  \neq\emptyset$ define
\[
\operatorname*{Rem}\left(  U,r\right)  \equiv\left\{
\begin{array}
[c]{c}%
head\left(  r\right)  :-\;body\left(  r\right)  :CS_{\operatorname*{Rem}%
}\left(  U,r\right)  ,\;Adv_{\operatorname*{Rem}}\left(  U,r\right)  \text{ if
}r\text{ is advancing}\\
head\left(  r\right)  :-\;body\left(  r\right)  :CS_{\operatorname*{Rem}%
}\left(  U,r\right)  ,\;Bool\left(  r\right)  \text{ if }r\text{ is
stationary}%
\end{array}
\right.
\]

\bigskip In other words, $\operatorname*{Rem}\left(  U,r\right)  $ is the part
of $r$ that contributes to the complement of $U$.

Define%
\[
\operatorname*{Rem}\left(  U,P\right)  \equiv\{\;\operatorname*{Rem}\left(
U,r\right)  :\;r\in Rules_{b}\left(  U,P\right)  \text{ and }%
CS_{\operatorname*{Rem}}\left(  U,r\right)  \neq\emptyset\;\}
\]

\bigskip That is $\operatorname*{Rem}\left(  U,P\right)  $ contain those parts
of the rules in $Rules_{b}\left(  U,P\right)  $ that contribute to the
complement of $U$.

Let $X\subseteq U$. For a rule $r$ define
\[
CS_{\epsilon}\left(  U,r,X\right)  \equiv\{\;(\mathbf{p}_{1}%
,\;...,\;\mathbf{p}_{n})\in CS\left(  r\right)  :
\]%
\[
\text{for }i=1,\;...,\;n\text{ }\{B_{i}^{+}\times\mathbf{p}_{i}\}\cap
U\subseteq X\text{ and }\{B_{i}^{-}\times\mathbf{p}_{i}\}\cap X=\emptyset\;\}
\]

That is $CS_{\epsilon}\left(  U,r,X\right)  $ is the set of those generalized
position tuples such that for them the "projection"\ of $body\left(  r\right)
$ onto $U$ is satisfied by $X$.

Finally
\[
\epsilon_{U}\left(  P,X\right)  \equiv\{
\]%
\[
r^{\prime}\equiv a:-\;B_{1}\backslash At\left(  U|_{\mathbf{p}_{1}}\right)
;\;...;\;B_{n}\backslash At\left(  U|_{\mathbf{p}_{n}}\right)
:\;\{(\mathbf{p}_{1},\;...,\;\mathbf{p}_{n})\},\;Q\text{ }|
\]%
\[
r\equiv a:-\;B_{1};\;...;\;B_{n}:O,Q\;\in\;\{\;r\in P:\;CS\left(  \epsilon
_{U},r,X\right)  \neq\emptyset\;\}\text{ and}%
\]%
\[
(\mathbf{p}_{1},\;...,\;\mathbf{p}_{n})\in CS_{\epsilon}\left(  U,r,X\right)
\;\}
\]

In other words, for every rule $r\in P$ such that$\;CS\left(  \epsilon
_{U},r,X\right)  \neq\emptyset$ and for every $(\mathbf{p}_{1}%
,\;...,\;\mathbf{p}_{n})\in CS_{\epsilon}\left(  U,r,X\right)  $ where the
"projection" of $body\left(  r\right)  $ onto $U$ is satisfied by $X$ at
$(\mathbf{p}_{1},\;...,\;\mathbf{p}_{n})$, we add to $\epsilon_{U}\left(
P,X\right)  $ a rule $r^{\prime}$, which is a part of rule $r$ that will be
active only for that $(\mathbf{p}_{1},\;...,\;\mathbf{p}_{n})$ with the
"projection"\ part removed.

\bigskip

\textbf{Theorem 1. }\emph{(The Splitting Set Theorem for Hybrid ASP). Let }%
$P$\emph{ be a Hybrid ASP\ program over }$Lit_{At}\times S$\emph{. Let
}$U\subseteq Lit_{At}\times S$\emph{ be a splitting set of }$P$\emph{ w.i.c.
}$J\subseteq S$\emph{. A set }$M$\emph{ is a answer set of }$P$\emph{ w.i.c.
}$J$\emph{ iff }$X\equiv M\cap U$\emph{ is a answer set of }$b_{U}\left(
P\right)  $\emph{ w.i.c. }$J$\emph{ and }$M\backslash U$\emph{ is a answer set
of }$\epsilon_{U}(P\backslash Rules_{b}\left(  U,P\right)  \cup
\operatorname*{Rem}\left(  U,P\right)  ,\;X)$\emph{ w.i.c. }$GP_{J}\left(
X\right)  $\emph{.}

\bigskip

\emph{Sketch of a proof. }We first prove that if $M$ is an answer set of $P$
w.i.c. $J$ then $X\equiv M\cap U$ is an answer set of $b_{U}\left(  P\right)
$ w.i.c. $J$. That is, we want to show that $X=%
{\displaystyle\bigcup\limits_{k=0}^{\infty}}
T\left[  b_{U}\left(  P\right)  ^{X,J},J\right]  ^{k}\left(  \emptyset\right)
$. In $\supseteq$ direction we show by induction on $k$ in one-step
provability operator $T\left[  b_{U}\left(  P\right)  ^{X,J},J\right]  ^{k}$
that if a rule $b_{U}\left(  r\right)  ^{X,J}$ in $b_{U}\left(  P\right)
^{X,J}$ derives $\left(  a,\mathbf{p}\right)  $ in $T\left[  b_{U}\left(
P\right)  ^{X,J},J\right]  ^{k+1}\left(  \emptyset\right)  $, then the rule
$r^{M,J}$ must derive $\left(  a,\mathbf{p}\right)  $ in $T\left[
P^{M,J},J\right]  ^{m+1}\left(  \emptyset\right)  $ for some $m$. In
$\subseteq$ direction we show by induction on $k$ in $T\left[  P^{M,J}%
,J\right]  ^{k}\left(  \emptyset\right)  $ that if $r^{M,J}$ derives $\left(
a,\mathbf{p}\right)  $ in $T\left[  P^{M,J},J\right]  ^{k+1}\left(
\emptyset\right)  $ where $\left(  a,\mathbf{p}\right)  \in U$, then
$b_{U}\left(  r\right)  ^{X,J}$ derives $\left(  a,\mathbf{p}\right)  $ in
$T\left[  b_{U}\left(  P\right)  ^{X,J},J\right]  ^{m+1}\left(  \emptyset
\right)  $ for some $m$.

We then proceed to prove that if $M$ is an answer set of $P$ w.i.c. $J$, and
$Y\equiv M\backslash U$ then $Y$ is an answer set of $Q\equiv\epsilon
_{U}(P\backslash Rules_{b}\left(  U,P\right)  \cup\operatorname*{Rem}\left(
U,P\right)  ,X)$\emph{ w.i.c. }$L\equiv GP_{J}\left(  X\right)  $. That is, we
want to show that $Y=%
{\displaystyle\bigcup\limits_{k=0}^{\infty}}
T\left[  Q^{Y,L},L\right]  ^{k}\left(  \emptyset\right)  $. In $\supseteq$
direction we prove by induction that if $r^{Y,L}$ derives $\left(
a,\mathbf{p}\right)  $ in $T\left[  Q^{Y,L},L\right]  ^{k+1}\left(
\emptyset\right)  $ then there is a corresponding rule $q^{M,J}$ in $P^{M,J}$
that derives $\left(  a,\mathbf{p}\right)  $ in $T\left[  P^{M,J},J\right]
^{m+1}\left(  \emptyset\right)  $ for some $m$. In $\subseteq$ direction we
prove by induction on $k$ in $T\left[  P^{M,J},J\right]  ^{k}\left(
\emptyset\right)  $ that if $q^{M,J}$ derives $\left(  a,\mathbf{p}\right)  $
in $T\left[  P^{M,J},J\right]  ^{k+1}\left(  \emptyset\right)  $ where
$\left(  a,\mathbf{p}\right)  \in M\backslash U$ then there is a corresponding
$r$ in $Q^{Y,L}$ that derives $\left(  a,\mathbf{p}\right)  $ in $T\left[
Q^{Y,L},L\right]  ^{m+1}\left(  \emptyset\right)  $ for some $m$.

To finish the proof we need to show that if $X\subseteq U$ is an answer set of
$b_{U}\left(  P\right)  $ w.i.c. $J$ and $Y\subseteq U^{C}$ is an answer set
of $Q$ w.i.c. $L$ then $M\equiv X\cup Y$ is an answer set of $P$ w.i.c. $J$.
That is we want to show that $M=%
{\displaystyle\bigcup\limits_{k=0}^{\infty}}
T\left[  P^{M,J},J\right]  ^{k}\left(  \emptyset\right)  $. We do so by
induction in both directions in a manner similar to the previous part of the
proof. $\square$

\bigskip

Similar to the Splitting Sequence Theorem of
\cite{DBLP:conf/iclp/LifschitzT94} we also prove the Splitting Sequence
Theorem for Hybrid ASP.

\textbf{Theorem 2.}\emph{ (The Splitting Sequence Theorem for Hybrid ASP). Let
}$\left\langle U_{\alpha}\right\rangle _{\alpha<\mu}$ \emph{be a monotone
continuous sequence of splitting sets for a Hybrid ASP\ program }$P$\emph{
over }$At\times S$\emph{ w.i.c. }$J\subseteq S$\emph{, and }$%
{\displaystyle\bigcup\limits_{\alpha<\mu}}
U_{\alpha}=Lit_{At}\times S$.\emph{ }$M$\emph{ is an answer set of }$P$\emph{
w.i.c. }$J$\emph{ iff }$M=%
{\displaystyle\bigcup\limits_{\alpha<\mu}}
X_{\alpha}$\emph{ for a sequence }$\left\langle X_{\alpha}\right\rangle
_{\alpha<\mu}$\emph{ s.t.}

\begin{itemize}
\item $X_{0}$\emph{ is an answer set of }$b_{U_{0}}\left(  P\right)  $\emph{
w.i.c. }$J$

\item \emph{for any }$\alpha$\emph{ such that }$\alpha+1<\mu$\emph{
}$X_{\alpha+1}$\emph{ is an answer set for }

$\epsilon_{U_{\alpha}}(b_{U_{\alpha+1}}\left(  P\right)  \;\backslash
\;Rules_{b}(U_{\alpha},b_{U_{\alpha+1}}\left(  P\right)  )\cup
\operatorname*{Rem}(U_{\alpha},b_{U_{\alpha+1}}\left(  P\right)  ),\;%
{\displaystyle\bigcup\limits_{\beta\leq\alpha}}
X_{\beta})$\emph{ w.i.c.}\emph{ }$L_{\alpha}\equiv GP_{J}(%
{\displaystyle\bigcup\limits_{\beta\leq\alpha}}
X_{\beta})$\emph{ and }$X_{\alpha+1}=M\cap(U_{\alpha+1}\backslash U_{\alpha}%
)$\emph{ and }$%
{\displaystyle\bigcup\limits_{\beta\leq\alpha}}
X_{\beta}$\emph{ is an answer set of }$b_{U_{\alpha}}\left(  P\right)  $\emph{
w.i.c. }$J$\emph{.}
\end{itemize}

\bigskip

The proof proceeds by the induction on $\alpha$ and is a direct application of
The Splitting Set Theorem for Hybrid ASP. 

In the Splitting Sequence Theorem for Hybrid ASP, $b_{U_{\alpha+1}}\left(
P\right)  $ is a program that derives $%
{\displaystyle\bigcup\limits_{\beta\leq\alpha+1}}
X_{\beta}$ as its answer set w.i.c. \ $J$. Now, $%
{\displaystyle\bigcup\limits_{\beta\leq\alpha+1}}
X_{\beta}$ $\subseteq%
{\displaystyle\bigcup\limits_{\beta\leq\alpha+1}}
U_{\beta}$. So, to derive $X_{\alpha+1}$ (i.e. the subset of $%
{\displaystyle\bigcup\limits_{\beta\leq\alpha+1}}
X_{\beta}$ that is in $U_{\alpha+1}\backslash U_{\alpha}$) we need to remove
from $b_{U_{\alpha+1}}\left(  P\right)  $ the rules that derive $%
{\displaystyle\bigcup\limits_{\beta\leq\alpha}}
X_{\beta}$. That is accomplished by subtracting from $b_{U_{\alpha+1}}\left(
P\right)  $ the rules $Rules_{b}(U_{\alpha},b_{U_{\alpha+1}}\left(  P\right)
)$. Nevertheless, this subtracts too much as some of the rules in
$Rules_{b}(U_{\alpha},b_{U_{\alpha+1}}\left(  P\right)  )$ contribute to
$X_{\alpha+1}$ for some generalized position tuples. The parts of those rules
that contribute to $X_{\alpha+1}$ are $\operatorname*{Rem}(U_{\alpha
},b_{U_{\alpha+1}}\left(  P\right)  )$, which we then add back. Applying
$\epsilon_{U_{\alpha}}$ operator to the resulting program (i.e. $b_{U_{\alpha
+1}}\left(  P\right)  \;\backslash\;Rules_{b}(U_{\alpha},b_{U_{\alpha+1}%
}\left(  P\right)  )\cup\operatorname*{Rem}(U_{\alpha},b_{U_{\alpha+1}}\left(
P\right)  )$) then removes the "useless"\ part of the rules with respect to $%
{\displaystyle\bigcup\limits_{\beta\leq\alpha}}
X_{\beta}$.

\bigskip

\section{An Application:\ Computing Answer Sets of Hybrid ASP\ Programs}

One of the applications of the Splitting Sequence Theorem for Hybrid ASP\ is
proving the correctness of a certain algorithm for computing answer sets of
certain types of Hybrid ASP\ programs. We will consider only the programs
where the set of generalized positions $S$ is such that if $\mathbf{p}\in S$
then $t\left(  \mathbf{p}\right)  =k\cdot\Delta t$ where $k\in%
\mathbb{N}
$, and for any advancing rule $r$ of any arity $n$, for any $(\mathbf{p}%
_{1},\;...,\;\mathbf{p}_{n})\in S^{n}$ we have that for all $\mathbf{q\in
}Adv\left(  r\right)  (\mathbf{p}_{1},...,\;\mathbf{p}_{n})$, $t\left(
\mathbf{q}\right)  =t\left(  \mathbf{p}_{n}\right)  +\Delta t$. That is, these
are the programs with generalized positions with discrete times of the form
$k\Delta t$, and whenever an advancing algorithm produces a new generalized
position, that generalized position has time larger by $\Delta t$ than the
largest time in the input arguments. All applications of Hybrid ASP known to
the author\ are restricted to such programs. This is the case for using Hybrid
ASP\ to diagnose failure of data processing pipelines, as described in
\cite{BomansonB19} and \cite{BrikX20}. It is the case for the Hybrid
ASP\ programs that are the result of translation from action languages Hybrid
AL \cite{BrikR17} and Hybrid ALE \cite{BomansonB19}. It is also the case for
using Hybrid ASP to compute optimal finite horizon policies in dynamic domains
\cite{HybridASP2}. 

\textbf{The algorithm.}

We will first describe the algorithm informally. We will use some of the new
notation which will be defined further below. The algorithm is based on the
observation that in Hybrid ASP the facts in the "future"\ cannot affect the
facts in the "past". That is for any two generalized position $\mathbf{p}$ and
$\mathbf{q}$, if $t\left(  \mathbf{p}\right)  <t\left(  \mathbf{q}\right)  $
then the state at $\mathbf{q}$ cannot be used to derive the state at
$\mathbf{p}$ (but the state at $\mathbf{p}$ can be used to derive the state at
$\mathbf{q}$). Consequently, it should be possible to first derive the states
at some minimal time $t_{\min}$, then derive the states at the time $t_{\min
}+\Delta t$, then derive the states at time $t_{\min}+2\Delta t$ and so on.

Without the loss of generality, we will assume that for any initial condition
$J\subseteq S$, there exists $\mathbf{p}\in J$ such that $t\left(
\mathbf{p}\right)  =0$. Let $P$ be a Hybrid ASP\ program over $Lit_{At}\times
S$. Let $J\subseteq S$ be an initial condition. The algorithm will be defined
inductively. Suppose the set $N$ of all the (literal, generalized
position)\ pairs for the generalized positions with time up to $k\cdot\Delta
t$ is derived by the algorithm for some $k$. The algorithm will first identify
all the advancing rules $Rules_{Adv}\left(  P,N,k\Delta t\right)  $ that could
derive generalized positions with time $\left(  k+1\right)  \cdot\Delta t$.
These are the advancing rules $r$ such that $N$ satisfies their body for some
$(\mathbf{p}_{1},...,\;\mathbf{p}_{n})\in CS\left(  r\right)  $, where
$n=arity\left(  r\right)  $ and the time of $\mathbf{p}_{n}$ is $k\cdot\Delta
t$. The set of the "next" generalized positions (i.e. the set of generalized
positions with time $\left(  k+1\right)  \cdot\Delta t$) is derived by
choosing a subset of the set of all the generalized positions derived by these
rules. To formally define such a choice of a subset we introduce a concept of
an \emph{advancing selector} $F$, which is a function s.t. for $M\subseteq
Lit_{At}\times S$ and $Z\subseteq S$, $F\left(  M,Z\right)  $ is a subset of
$Z$. We will denote the set of "next" generalized positions derived in this
manner by $NextGP\left(  P,F,N,k\Delta t\right)  $.

Now, for every "next" generalized position $\mathbf{q}$ in $NextGP\left(
P,F,N,k\Delta t\right)  $ derived by an advancing rule $r\in Rules_{Adv}%
\left(  P,N,k\Delta t\right)  $, it must be that $(head\left(  r\right)
,\;\mathbf{q})$ is derived. So, for every $\mathbf{q}$ there is a set of
literals that will be derived at $\mathbf{q}$ by the advancing rules in
$Rules_{Adv}\left(  P,N,k\Delta t\right)  $. This set of literals will be
denoted by $Head_{Adv}(P,N,\mathbf{q})$.

Next we turn our attention to the role of the stationary rules in deriving
hybrid state at a "next" generalized position $\mathbf{q}$. There is a set of
stationary rules that can contribute to the hybrid state at $\mathbf{q}$. If
such a stationary rule $r$ has $n$ blocks, then the first $n-1$ blocks are
satisfied by $N$ (at some generalized positions $\mathbf{p}_{1}%
,\;...,\;\mathbf{p}_{n-1}$) and $(\mathbf{p}_{1},\;...,\;\mathbf{p}%
_{n-1},\;\mathbf{q})$ are in $CS\left(  r\right)  \cap Bool\left(  r\right)
$. Thus, only the last block, which we will denote by $B_{n}$ needs to be
evaluated at $\mathbf{q}$. Thus, the relevant part of such a stationary rule
$r$ is a regular ASP rule of the form $head\left(  r\right)  :-B_{n}$. All
such regular ASP\ rules applicable at $\mathbf{q}$ will be denoted by
$\operatorname*{Red}\nolimits_{App}(P,\ N,\;\mathbf{q})$. A state at
$\mathbf{q}$ is then an answer set of a regular ASP\ program
$\operatorname*{Red}\nolimits_{App}(P,\ N,\;\mathbf{q})\cup\{[h:-]\;:\;h\in
Head_{Adv}(P,N,\mathbf{q})\}$. To formally define such a choice we will use a
concept of a \emph{stationary selector }$D$, which we will define further below.

We will now define the algorithm formally.

For a set $N\subseteq Lit_{At}\times S$ and generalized positions $\mathbf{p}$
and $\mathbf{q}$, let
\[
Rules_{Adv}\left(  P,N,k\Delta t\right)  \equiv\{r\in P\;:\;r\text{ is an
advancing rule and there is }%
\]%
\[
\text{ }(\mathbf{p}_{1},\;...,\;\mathbf{p}_{n})\in GP_{J}\left(  N\right)
^{n}\cap CS\left(  r\right)  \text{ with }t\left(  \mathbf{p}_{n}\right)
=k\cdot\Delta t\text{ and }N\models_{J}(body\left(  r\right)  ,\;(\mathbf{p}%
_{1},...,\;\mathbf{p}_{n})\}
\]

Let $\mathbf{p}_{1},\;...,\;\mathbf{p}_{n}\in GP_{J}\left(  N\right)  $. We
define the set of advancing rules active at $\mathbf{p}_{1},\;...,\;\mathbf{p}%
_{n}$ relative to $N$ as%
\[
Rules_{Adv}(P,\;N,\;(\mathbf{p}_{1},\;...,\;\mathbf{p}_{n}))\equiv\{r\in
Rules_{Adv}\left(  P,\;N,\;t\left(  \mathbf{p}_{n}\right)  \right)  \;:\text{
}(\mathbf{p}_{1},\;...,\;\mathbf{p}_{n})\in CS\left(  r\right)  \}.
\]
That is, $Rules_{Adv}(P,\;N,\;(\mathbf{p}_{1},\;...,\;\mathbf{p}_{n}))$ is the
set of the advancing rules whose body is satisfied by $N$ at $(\mathbf{p}%
_{1},\;...,\;\mathbf{p}_{n})$ and $(\mathbf{p}_{1},\;...,\;\mathbf{p}_{n})\in
CS\left(  r\right)  $.

We define the set of "next" generalized positions at $\mathbf{p}%
_{1},\;...,\;\mathbf{p}_{n}$ relative to $N$ as%
\[
NextGP(P,N,\;(\mathbf{p}_{1},\;...,\;\mathbf{p}_{n}))\equiv%
{\displaystyle\bigcup\limits_{r\in Rules_{Adv}(P,\;N,\;(\mathbf{p}%
_{1},\;...,\;\mathbf{p}_{n}))}}
Adv\left(  r\right)  (\mathbf{p}_{1},\;...,\;\mathbf{p}_{n}).
\]
That is $NextGP(P,N,\;(\mathbf{p}_{1},\;...,\;\mathbf{p}_{n}))$ is the set of
"next" generalized positions generated by any advancing rule active at
$\mathbf{p}_{1},\;...,\;\mathbf{p}_{n}$ relative to $N$.

For a time $k\cdot\Delta t$, we define the set of all the "next" generalized
positions relative to $N$, $k\cdot\Delta t$ and an advancing selector $F$ as%
\[
NextGP(P,F,N,k\Delta t)\equiv F(N,\;%
{\displaystyle\bigcup\limits_{\substack{_{\substack{n\geq1\\\mathbf{p}%
_{1},\;...,\;\mathbf{p}_{n}\in GP_{J}\left(  N\right)  }}\\t\left(
\mathbf{p}_{n}\right)  =k\Delta t}}}
NextGP(P,N,\;(\mathbf{p}_{1},\;...,\;\mathbf{p}_{n}))\;).
\]

The set of all heads at $\mathbf{q}\in NextGP\left(  P,F,N,k\Delta t\right)  $
relative to $N$ is then%
\[
Head_{Adv}(P,\;N,\;\mathbf{q})\equiv\{\;head\left(  r\right)  \;:\text{there
exists }\mathbf{p}_{1},\;...,\;\mathbf{p}_{n}\in GP_{J}\left(  N\right)
\text{ and}%
\]%
\[
\text{ }r\in Rules_{Adv}(P,\;N,\;(\mathbf{p}_{1},\;...,\mathbf{\;p}_{n}))
\]%
\[
\text{such that }\mathbf{q}\in Adv\left(  r\right)  (\mathbf{p}_{1}%
,\;...,\;\mathbf{p}_{n})\}\text{.}%
\]

Let $\mathbf{p}_{1},\;...,\;\mathbf{p}_{n}\in GP_{J}\left(  N\right)  $. We
define the set of stationary rules active at $\mathbf{p}_{1}%
,\;...,\;\mathbf{p}_{n}$ relative to $N$ as%
\[
Rules_{Stat}(P,\;N,\;(\mathbf{p}_{1},\;...,\;\mathbf{p}_{n}))\equiv\{r\in
P\;:r\text{ is stationary and}%
\]%
\[
\text{ }(\mathbf{p}_{1},\;...,\;\mathbf{p}_{n})\in CS\left(  r\right)  \cap
Bool\left(  r\right)  \text{ and for }i=1,\;...,\;n-1\text{ }N\models
_{J}(B_{i},\;\mathbf{p}_{i})\;\}\text{.}%
\]
That is $Rules_{Stat}(P,\;N,\;(\mathbf{p}_{1},\;...,\;\mathbf{p}_{n}))$ is the
set of stationary rules with $n-1$ blocks satisfied by $N$ at $\mathbf{p}%
_{1},\;...,\;\mathbf{p}_{n-1}$ respectively, and $(\mathbf{p}_{1}%
,\;...,\;\mathbf{p}_{n})\in CS\left(  r\right)  \cap Bool\left(  r\right)  $.

We define a \emph{stationary selector }$D$ to be a function such that for
$M\subseteq At\times S$ for $\mathbf{z}\in S$ for an ASP\ program $U$,
$D(M,\;\mathbf{z},\;U)$ is an answer set of $U$. That is, a stationary
selector chooses one of answer sets of a regular ASP\ programs $U$.

For a stationary rule $r$ of the form $a:-B_{1};\;...;\;B_{n}:O,H\,$, we
define an \emph{applicable reduct} of $r$
\[
\operatorname*{Red}\nolimits_{App}\left(  r\right)  \equiv\{a:-\;B_{n}%
\}\text{.}%
\]

For $\mathbf{z}\in NewGP(P,F,N,k\Delta t)$ we define the active reduct of $P$
at $\mathbf{z}$ relative to $N$ as%
\[
\operatorname*{Red}\nolimits_{App}(P,\;N,\;\mathbf{z})\equiv
\{\operatorname*{Red}\nolimits_{App}\left(  r\right)  :\;\text{there exists
}n\geq1\text{ and }(\mathbf{p}_{1},\;...,\;\mathbf{p}_{n-1})\in GP_{J}\left(
N\right)  ^{n-1}%
\]%
\[
\text{ such that }r\in Rules_{Stat}(P,\;N,\;(\mathbf{p}_{1},\;...,\;\mathbf{p}%
_{n-1},\;\mathbf{z})\;\}
\]
\bigskip

Finally, for $N\subseteq At\times S$ and $i\in%
\mathbb{N}
$ let $N\left[  i\right]  \equiv\{(a,\;\mathbf{p})\in N:\;t\left(
\mathbf{p}\right)  =i\cdot\Delta t\}$. Similarly for $Z\subseteq S$,
$Z[i]\equiv\{\mathbf{p}\in Z\;:\;t\left(  \mathbf{p}\right)  =i\cdot\Delta
t\}$.

We are now ready to formally specify our algorithm. We define a sequence of
sets $\left\langle Y_{i}\right\rangle _{i\geq0}$ , $Y_{i}\subseteq(At\times
S)[i]$ as follows:%
\[
Y_{0}\equiv%
{\displaystyle\bigcup\limits_{\mathbf{z}\in J\left[  0\right]  }}
D(\emptyset,\;\mathbf{z},\;\operatorname*{Red}\nolimits_{App}(P,\;\emptyset
,\;\mathbf{z))\times z}%
\]

That is, the state at any generalized position $\mathbf{z}\in J$ with time
equal to $0$ is determined by taking all the stationary rules $r$ with one
block (i.e. rules of the form $a:-B:O,H$ ) such that $\mathbf{z}\in O\cap H$,
composing a regular ASP\ program from the reducts of the form $a:-B$ derived
from those rules, and then finding an answer set of that program.

Now, suppose $Y_{i}$ are defined for $0\leq i\leq k$ and $Y_{k}\neq\emptyset$.
Let%
\[
Z_{k+1}\equiv NextGP(P,F,\;%
{\displaystyle\bigcup\limits_{i=0}^{k}}
Y_{i},\;k\Delta t).
\]
That is $Z_{k+1}$ is the set of generalized positions with time $\left(
k+1\right)  \Delta t$ derived by the advancing rules $Rules_{Adv}\left(  P,\;%
{\displaystyle\bigcup\limits_{i=0}^{k}}
Y_{i},\;k\Delta t\right)  $.

Let%
\[
Y_{k+1}\equiv%
{\displaystyle\bigcup\limits_{\mathbf{z}\in Z_{k+1}}}
D(%
{\displaystyle\bigcup\limits_{i=0}^{k}}
Y_{i},\;\mathbf{z},\;\operatorname*{Red}\nolimits_{App}(P,\;%
{\displaystyle\bigcup\limits_{i=0}^{k}}
Y_{i},\;\mathbf{z})\cup
\]%
\[
\{[a:-]\;:\;a\in Head_{Adv}(P,\;%
{\displaystyle\bigcup\limits_{i=0}^{k}}
Y_{i},\;\mathbf{z})\})\;\times\;\mathbf{z}%
\]
if $D(...)\neq\emptyset$ and $Y_{k+1}\equiv\emptyset$ otherwise.

That is, $Y_{k+1}$ is a collection of hybrid states $\left(  Y_{k+1}%
|_{\mathbf{z}},\mathbf{z}\right)  $ where $\mathbf{z\in Z}_{k+1}$, and where
$Y_{k+1}|_{\mathbf{z}}$ is an answer set of a regular ASP\ program composed of
the active reducts of the stationary rules that can contribute to $\mathbf{z}$
and the heads of the advancing rules that derive $\mathbf{z}$.\newline

\textbf{Theorem 3.} $M$ is an answer set of $P$ w.i.c. $J$ iff there is
advancing selector $F$ and a stationary selector $D$ such that $%
{\displaystyle\bigcup\limits_{i=0}^{\infty}}
Y_{i}=M$ with $F$ and $D$.

\bigskip

\emph{Sketch of a proof.} We begin by specifying a sequence of splitting sets
$\left\langle U_{i}\right\rangle _{i=0}^{\infty}$ defined as%
\[
U_{i}=Lit_{At}\times\{\mathbf{p}:\;\mathbf{p}\in S\text{ and }0\leq t\left(
\mathbf{p}\right)  \leq i\Delta t\;\}
\]

We then first show that $Y_{0}$ is an answer set of $b_{U_{0}}\left(
P\right)  $ w.i.c. $J$. The rules that can contribute to $Y_{0}$ are
stationary-1 rules $r$ such as $CS\left(  r\right)  \cap Bool\left(  r\right)
\cap J\left[  0\right]  \neq\emptyset$. These rules will contribute regular
ASP\ rules to $\operatorname*{Red}\nolimits_{App}(P,\;\emptyset,\;\mathbf{z})$
for every $\mathbf{z}\in J\left[  0\right]  $. We then show that
$D(\emptyset,\;\mathbf{z},\;\operatorname*{Red}\nolimits_{App}(P,\emptyset
,\mathbf{z}))$ is an answer set of $\operatorname*{Red}\nolimits_{App}%
(P,\emptyset,\mathbf{z})$ iff $D(\emptyset,\;\mathbf{z},\;\operatorname*{Red}%
\nolimits_{App}(P,\emptyset,\mathbf{z}))\times\mathbf{z}$ is an answer set of
$b_{U_{0}}\left(  P\right)  $ w.i.c. $J$. 

The rest is proven by induction using The Splitting Sequence Theorem. That is
$M\left[  k+1\right]  $ is an answer set of $E=\epsilon_{U_{k}}(b_{U_{k+1}%
}\left(  P\right)  \backslash Rules_{b}(U_{k},b_{U_{k+1}}\left(  P\right)
)\cup\operatorname*{Rem}(U_{k},b_{U_{k+1}}\left(  P\right)  ),\;%
{\displaystyle\bigcup\limits_{i\leq k}}
M\left[  i\right]  )$ w.i.c. $GP_{J}(L)$, where $L=%
{\displaystyle\bigcup\limits_{i\leq k}}
M\left[  i\right]  $ iff there exists an advancing selector $F$ and a
stationary selector $D$ such that $M\left[  k+1\right]  $ is equal to
$Y_{k+1}$ as defined by the algorithm.

\bigskip For the forward direction of the inductive step we define
$F(N,\;Y)\equiv Y\cap GP\left(  M\right)  $. We define
\[
D(N,\;\mathbf{p},\;Q)\equiv\left\{
\begin{array}
[c]{c}%
At\left(  N|_{\mathbf{p}}\right)  \text{ if }At\left(  N|_{\mathbf{p}}\right)
\text{ is an answer set of }Q\\
\emptyset\text{ otherwise}%
\end{array}
\right.
\]

We then show $GP\left(  M\left[  k+1\right]  \right)  =NextGP(P,F,L,k\Delta
t)$. We then use the induction on one-step provability operator\ $T\left[
E^{M\left[  k+1\right]  ,\;GP_{J}\left(  L\right)  },GP_{J}\left(  L\right)
\right]  ^{j}$ to show that if $M\left[  k+1\right]  $ is an answer set of $E$
w.i.c. $GP_{J}\left(  L\right)  $ then $M\left[  k+1\right]  |_{\mathbf{p}%
}=Y_{k+1}|_{\mathbf{p}}$. That is we show that if $M\left[  k+1\right]  $ is
an answer set of $E$ w.i.c. $GP_{J}\left(  L\right)  $ then the algorithm
derives it as $Y_{k+1}$.

For the reverse direction we first show $\{(head\left(  r\right)
,\;\mathbf{p}):\;r\in Head_{Adv}(P,\;L,\;\mathbf{p}),\;\mathbf{p}\in GP\left(
Y_{k}\right)  \}\subseteq T\left[  E^{Y_{k+1},GP_{J}\left(  L\right)  }%
,GP_{J}\left(  L\right)  \right]  ^{1}\left(  \emptyset\right)  $. That is we
show that the literals of $Head_{Adv}(P,\;L,\;\mathbf{p})$ are also derived by
$E$ at $\mathbf{p}$. We then use induction on one step provability $T\left[
K^{At\left(  Y_{k+1}|_{\mathbf{p}}\right)  }\right]  ^{i}$, where
$\ K\equiv\operatorname*{Red}\nolimits_{App}(P,L,\mathbf{p})$ to show that for
all $\mathbf{p}\in GP\left(  Y_{k+1}\right)  $ it is the case that%
\ $%
{\displaystyle\bigcup\limits_{i\geq0}}
T\left[  K^{At\left(  Y_{k+1}|_{\mathbf{p}}\right)  }\right]  ^{i}\left(
\emptyset\right)  \times\mathbf{p}\subseteq%
{\displaystyle\bigcup\limits_{j\geq0}}
T\left[  E^{Y_{k+1},GP_{J}\left(  L\right)  },GP_{J}\left(  L\right)  \right]
^{j}\left(  \emptyset\right)  $, for some $j$. That is, we show that the
literals derived by the regular ASP\ program $\operatorname*{Red}%
\nolimits_{App}(P,L,\mathbf{p})$ are also derived by $E$ at $\mathbf{p}$. But
this merely shows that $Y\equiv%
{\displaystyle\bigcup\limits_{i=0}^{\infty}}
Y_{i}$ $\subseteq$ $%
{\displaystyle\bigcup\limits_{j\geq0}}
T[P^{Y,J},J]\left(  \emptyset\right)  $. We also need to show that $%
{\displaystyle\bigcup\limits_{j\geq0}}
T[P^{Y,J},J]\left(  \emptyset\right)  \subseteq Y$.

We do that by using induction on one step provability operator $T\left[
E^{Y_{k+1},GP_{J}\left(  L\right)  },GP_{J}\left(  L\right)  \right]  ^{j}$ to
show that for all $\mathbf{p}\in GP\left(  Y_{k+1}\right)  $ it is the case
that $%
{\displaystyle\bigcup\limits_{j\geq0}}
T\left[  E^{Y_{k+1},GP_{J}\left(  L\right)  },GP_{J}\left(  L\right)  \right]
^{j}\left(  \emptyset\right)  $ is a subset of $%
{\displaystyle\bigcup\limits_{i\geq0}}
T\left[  K^{At\left(  Y_{k+1}|_{\mathbf{p}}\right)  }\right]  ^{i}\left(
\emptyset\right)  \times\mathbf{p}$.

This completes the proof of the theorem. $\square$

\bigskip The algorithm computes an answer set of the Hybrid ASP program $P$
w.i.c. $J$ inductively, by computing a subset of the answer set at time $0$,
then at time $\Delta t$, and so on through time $k\Delta t$. Moreover, the
aglorithm reduces the process of computing an answer set of a Hybrid ASP
program to the repeated application of two processes:\ the process of
computing the set of "next"\ generalized positions, and the process of
computing an answer set of a regular ASP\ program derived from advancing and
stationary Hybrid ASP\ rules applicable at these "next" generalized positions.

It's worth noting that the algorithm is a more general form of The Local
Algorithm \cite{HybridASP2}, variation of which is also discussed in
\cite{BomansonB19}.

\section{Conclusion}

The paper presents The Splitting Set Theorem for Hybrid ASP, which is\ the
equivalent for Hybrid ASP\ of the Splitting Set Theorem
\cite{DBLP:conf/iclp/LifschitzT94}, and the Splitting Sequence Theorem for
Hybrid ASP (which is the equivalent for Hybrid ASP\ of The Splitting Sequence
Theorem). The original Splitting Set Theorem proved to be a widely used
result. It is the author's hope that the new theorem will likewise prove to
have many applications. The paper discusses one of the applications of the
theorems to computing answer sets of Hybrid ASP\ programs.\newline%
\newline\textbf{Acknowledgements.} The author would like to thank Jori
Bomanson for insightful comments that helped to enhance the paper.

\bibliographystyle{eptcs}
\bibliography{bibliography}

\end{document}